\documentclass[letterpaper, 10 pt, conference]{ieeeconf}  % Comment this line out if you need a4paper
\pdfoutput=1

\IEEEoverridecommandlockouts                              % This command is only needed if 
\usepackage{listings}
\usepackage{bm}
\usepackage{amsmath} % assumes amsmath package installed
\usepackage{amssymb}
\usepackage{graphicx}
\usepackage{tikz}
\usepackage{pgfplots}
\usetikzlibrary{positioning}
\usetikzlibrary{shadows}
\usetikzlibrary{shapes.multipart}
\usetikzlibrary{fit}
\usepgfplotslibrary{groupplots}
\usepackage{xcolor}
\usepackage{booktabs}
\usepackage{soul}

\definecolor{codegreen}{rgb}{0,0.6,0}
\definecolor{codegray}{rgb}{0.5,0.5,0.5}
\definecolor{codepurple}{rgb}{0.58,0,0.82}
\definecolor{backcolour}{rgb}{0.95,0.95,0.92}
\lstdefinestyle{mystyle}{
	backgroundcolor=\color{backcolour}, commentstyle=\color{codegreen},
	keywordstyle=\color{magenta},
	numberstyle=\tiny\color{codegray},
	stringstyle=\color{codepurple},
	basicstyle=\ttfamily\footnotesize,
	breakatwhitespace=false,         
	breaklines=true,                 
	captionpos=b,                    
	keepspaces=true,                 
	numbers=left,                    
	numbersep=5pt,                  
	showspaces=false,                
	showstringspaces=false,
	showtabs=false,                  
	tabsize=2
}
\newcommand{\de}{\mathrm{d}}
\newcommand{\yvec}{\mathbf{y}}

\newcommand{\avec}{\mathbf{a}}
\newcommand{\cvec}{\mathbf{c}}
\newcommand{\zvec}{\mathbf{z}}
\newcommand{\xvec}{\mathbf{x}}

\newcommand{\epsilonvec}{\bm{\epsilon}}
\newcommand{\omegavec}{\bm{\omega}}

\title{\LARGE \bf
Deep Probabilistic Movement Primitives with a Bayesian Aggregator 
}

\author{Michael Przystupa$^{1}$, Faezeh Haghverd$^{1}$, Martin Jagersand$^{1}$, Samuele Tosatto$^{2}$% <-this % stops a space
\thanks{$^{1}$ University of Alberta, Canada. 
        \{{\tt przystup}, {\tt haghverd}, {\tt mj7}\}{@ualberta.ca}}%
\thanks{$^{2}$ University of Innsbruck, Austria.
        {\tt samuele.tosatto}{@uibk.ac.at}}%
}

\begin{document}

\maketitle
\thispagestyle{empty}
\pagestyle{empty}

%%%%%%%%%%%%%%%%%%%%%%%%%%%%%%%%%%%%%%%%%%%%%%%%%%%%%%%%%%%%%%%%%%%%%%%%%%%%%%%%
\begin{abstract}
Movement primitives are trainable parametric models that reproduce robotic movements starting from a limited set of demonstrations. Previous works proposed simple linear models that exhibited high sample efficiency and generalization power by allowing \textsl{temporal modulation} of movements (reproducing movements faster or slower), \textsl{blending} (merging two movements into one), \textsl{via-point conditioning} (constraining a movement to meet some particular via-points) and \textsl{context conditioning} (generation of movements based on an observed variable, e.g., position of an object). Previous works have proposed neural network-based motor primitive models, having demonstrated their capacity to perform tasks with some forms of input conditioning or time-modulation representations.  However, there has not been a single unified deep movement primitive's model proposed that is capable of all previous operations, limiting neural movement primitive's potential applications. This paper proposes a deep movement primitive architecture that encodes all the operations above and uses a Bayesian context aggregator that allows a more sound context conditioning and blending. Our results demonstrate our approach can scale to reproduce complex motions on a larger variety of input choices compared to baselines while maintaining operations of linear movement primitives provide. 
\end{abstract}

\section{Introduction}
\label{sec:introduction}
Learning from demonstration (LfD) is a technique that enables robots to acquire complex and adaptive skills by observing and reproducing human demonstrations \cite{argall_survey_2009,schaal_is_1999,ravichandar_recent_2020,zhou_learning_2019}. A key challenge in LfD is how to represent and generalize demonstrated motions in a robust and flexible way. Movement primitives (MPs) are a popular approach to address this challenge. MPs are parametric models that capture the essential features of a motion while allowing for variations and modifications according to different situations. The possibility of modifying the motion according to different needs is crucial to obtain an efficient mechanism to extrapolate new movements from scarce data. 

This generalization has previously been achieved by using probabilistic models to represent motor skills. Treating motor skills as probabilistic models captures the uncertainty in demonstration data that deterministic models would otherwise assume are fixed. Probabilistic models further enable  \textsl{variable conditioning} in which predictions can be refined through new observations. These allow movement primitives to adapt to new information during a motion.

\begin{figure}
\centering
\input{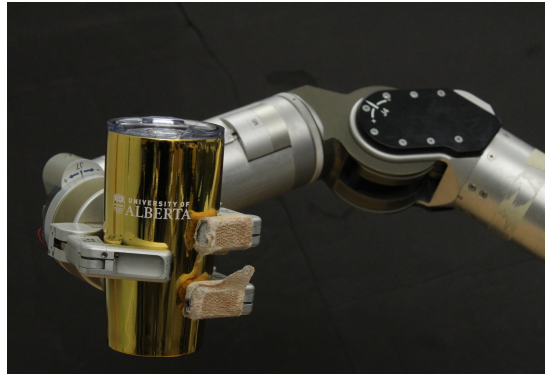}
\caption{Barret WAM preparing to perform rhythmic motions encoded with our deep probabilistic movement primitives to shake a Mojito.}
%\input{figures/kinova_fig_1_try.tex}
%\caption{Pick and place task with Kinova Gen 3 light experiments. DeepProMP can use context points to specify a task without via-point information}

\label{fig:shaking-wam}
\end{figure}

Probabilistic movement primitives (ProMPs) were one of the first frameworks to use probabilistic modeling to learn motor skills from human demonstrations. ProMPs are composed of a linear-Gaussian model, providing high mathematical tractability and allowing various operations, such as \textsl{via-point} and \textsl{context conditioning}, \textsl{blending}, \textsl{temporal modulation}, and \textsl{rhythmic movements} \cite{paraschos_probabilistic_2013,paraschos_using_2018}. Via-point conditioning allows sampling motions that pass through some desired robot configurations, whereas context conditioning generates motions based on some external variables (e.g., the position of an object in the scene). Blending generates an unseen movement by interpolating two learned ones. Temporal modulation allows the modification of the velocity profile of a given movement. Rhythmic movements enable a continuous repetition of the learned movement. 

However, Gaussian-linear models are limited to unimodal distributions and cannot represent rich, multimodal datasets or deal with high-dimensional variables like images \cite{colome_dimensionality_2018,tosatto_contextual_2020}. Recent approaches overcome these limitations with deep learning techniques. Some researchers have proposed autoregressive neural-networks-based movement primitives \cite{kipf_compile_2019,sharma_directed-info_2018,shankar_learning_2020,gupta_relay_2020,noseworthy_task-conditioned_2020}. These models are \textsl{time-discrete}, which leads to performance degradation with high frequencies \cite{bahl_neural_2020}. Other research proposes continuous-time approaches that use a deep model to generate the parameters of movement primitive models \cite{bahl_neural_2020,bahl_hierarchical_2021}, but the fundamental movements are still implemented with a linear model. 

% Further work has extended continuous-time to end-to-end deep learning approaches using the Conditional Neural Processes (CNPs) \cite{garnelo_conditional_2018,garnelo_neural_2018} which solves \textsl{meta-learning} problems \cite{wang2016learning,reed2017few}. CNPs encode input-output pairs through mean-aggregation to define context latent variable $\zvec$, where works have considered both deterministic and stochastic variations. This context variable implicitly defines a function class that can be used to make predictions of the function values when given new inputs (query points). 

The pivotal work of Seker et al. \cite{seker_conditional_2019} overcomes the limitations of previous work by proposing a continuous-time, non-linear representation of motor skills. These motor primitives utilizes Conditional Neural Processes (CNPs) \cite{garnelo_conditional_2018,garnelo_neural_2018}, an encoder-decoder model that allows aggregating multiple input-output pairs to form a latent representation of a function. In their framework, called Conditional Neural Movement Primitives (CNMPs), Seker et al. utilize time as the input variable and joint configurations as the output of CNPs. The aggregation of multiple time-joint-configuration pairs forms a latent representation of the desired motion. The latent representation is then used by a non-linear model to produce the continuous-time association between the two variables.  %utilizes this framework to propose Conditional Movement Primitives (CNMPS). The intuition is that a motor skill can be viewed as function mapping inputs (time or phase) to an output (joint configuration or end-effector's pose). When learning from a set of demonstrated motor skills, a movement primitive framework should learn to perform a variety of motor skills. When we query a movement primitive framework with a set of via-points (input-output pairs) or context variables, the  model predicts a - potentially - unseen motor skill. CNMPs enable via-point and context conditioning and have been tested on several real-robotic tasks. They are a clever application of deep-learning techniques because of their flexibility in learning associations between high-dimensional data and motor skills.

 However, CNMPs have several limitations: 1) CNMPs use a deterministic latent variable $\zvec$. The stochasticity is only provided by a Gaussian output, which is well suited to capture a unimodal \textsl{aleatoric} uncertainty; but cannot represent multimodal distributions. 2) The authors do not discuss how to provide motion blending, temporal modulation, and rhythmic movements. 3) CNMPs can still produce trajectories that are far from the desired via-points.% and assume dependence between context variables and via-points as inputs. %and have not been tested to work with high-dimensional inputs such as images.  

In this paper, we aim to mitigate these weaknesses and propose a deep probabilistic movement framework that exposes all the operations of classic ProMPs while enabling working with high-dimensional variables and maintaining high precision. Our deep probabilistic movement primitives (DeepProMPs) replace the CNMP deterministic mean aggregator with a Bayesian aggregator (BA) which has been shown to have superior predictive performance \cite{volpp_bayesian_2020}. We demonstrate this formulation provides a natural means to perform motion blending directly in the latent space. We assume our latent variable is independent when conditioned on either via-points or context variables. This enables flexibility when predicting motor skills and can be learned via auxiliary task learning. We further show how to implement both temporal modulation and rhythmic movement while maintaining high accuracy through state-of-the-art machine learning techniques, such as iterative optimization \cite{xia2023GANinversion} to improve via-point conditioning accuracy, and utilization input representations similar to positional encoding \cite{vaswani_attention_2017}. %Our motor primitive model uses techniques from other areas of machine learning like positional encoding \cite{vaswani_attention_2017}, iterative refinement \cite{xia2023GANinversion}, auxiliary task learning \cite{garnelo_conditional_2018}, and variational inference \cite{kingma_stochastic_2014} to achieve these properties, which will be explained better in the corpus of this paper.

In summary, our paper proposes a complete probabilistic deep-learning extension to ProMPs. Our model allows via-point and context conditioning, motion blending, temporal modulation, and rhythmic movements. Unlike ProMPs, our model can work with multimodal datasets and high-dimensional inputs. We exhaustively compare our model with ProMPs and CNMPs in virtual environments and on real robots.  Our results suggest that compared to CNMP alternatives, our model is better at reconstructing demonstrations across variations of specified input data, including high-dimensional data like images. Even when CNMP variants are trained to mimic these aspects, we find our model is superior in motor skill prediction. %We also demonstrate that our model extends easily to complex data types like images enabling more flexibility that ProMPs when conditioning on context information.

\section{Problem Statement}
\label{sec:background}

%MPs are parametric models that map a (normalized) time variable $\xvec$ to a robotic joint configuration $\yvec \in \mathbb{R}^d$, i.e., $\yvec = f_\zvec(\xvec) + \epsilon$ where $\zvec$ is a vector of parameters encoding the movement and $\epsilon$ is an independent and identically distributed (i.i.d.) noise. 
In this paper, we consider the problem of learning robotic movements from human demonstrations. A human demonstration can be summarized as a set of via-points $\mathcal{A} = \{\avec_i = (\xvec_i, \yvec_i)\}_{i=1}^{n}$, where via-points $\avec_i$ indicate the configuration $\yvec_i \in \mathbb{R}^d$ of the robot at the normalized time $\xvec_i$\footnote{We denote the time variable $\xvec$ with bold characters since rhythmic movements (Section~\ref{sec:deep-promps}) have a time variable that is formed by two components.}. Furthermore, each demonstration is accompanied by a variable number $m$ of \textsl{context variables} $\mathcal{C} = \{\cvec_i\}_{i=1}^{m}$ that describe some features of the environment (e.g., the pose of an object in the scene) or some desired goals (e.g., the amount of liquid to be poured in a glass). MPs aim to learn the relation between the time variable $\xvec$ and the joint configuration $\yvec$ by imitating the user's behavior in relation to the context variables $\cvec$.
ProMPs provide a full probabilistic view on the matter where $\zvec, \yvec$ and $\cvec$ are treated as stochastic variables \cite{paraschos_probabilistic_2013,paraschos_using_2018}. The probabilistic model allows both to sample movements similar to the one provided by the demonstration and to condition movements based on via-points and context variables, i.e., $p(\yvec | \xvec, \mathcal{A}, \mathcal{C}) = \int p(\yvec | \zvec, \xvec) p(\zvec | \mathcal{A}, \mathcal{C}) \de \zvec$. Usually, the number of conditioning variables $n$ and $m$ during deployment is much lower than the one in the dataset. ProMPs implement a linear-Gaussian model that allows mathematical tractability to solve the integrals in the closed form at the cost of lower expressivity. Due to their simple structure, ProMPs cannot represent complex data distributions. In the state-of-the-art literature, this limitation is overcome by using deep neural networks. 

\subsection{Conditional Neural Movement Primitives}
%CNMPS also can include context variables $\cvec_i$ as inputs to $\phi$.
CNMPs aim to provide a neural-network architecture to represent movement primitives and allow via-point and context conditioning. In CMNPs, each trajectory is associated with a latent variable $\zvec$ that represents the motion. CNMPs are, therefore, composed of two main modules: a deterministic mean aggregation encoder $\zvec = \frac{1}{n}(\sum_{i=1}^n \phi(\avec_i, \cvec_i)$, where $\phi$ is a neural network, and a movement generator (decoder) $p(\yvec | \zvec, \xvec)$.  The encoder associates a set of via-points $\avec_1, \dots, \avec_n$ and context variables $\cvec_1, \dots, \cvec_m$ with a deterministic latent representation of the movement $\zvec$. The encoder takes as input the time variable $\xvec$ and the latent motion representation $\zvec$ and outputs a Gaussian distribution over robot configurations $\yvec$. CNMPs can be trained with classic gradient descent techniques and allow conditioning with high-dimensional variables (e.g., $\{\cvec_i\}$ can be images). However, they can only learn unimodal uncertainty since $\yvec$ is Gaussian. 

\subsection{Bayesian Context Aggregator}\label{sec:background-bca}
Bayesian aggregation is a probabilistic aggregation technique that addresses the mean aggregator's inability to represent uncertainty. For example, the latent variable $\zvec$ should have a higher variance when only a few samples are provided, while it should have less variance when more samples are provided. With the mean aggregator, CNMPs postpone the problem of quantifying the uncertainty to the last layer with the neural network, which is unaware of how many conditioning variables have been presented. Therefore it cannot make a good prediction of the epistemic uncertainty. The Bayesian aggregator overcomes this issue by applying Bayesian inference directly to the latent variable.

We assume a prior distribution $p_0(\zvec) = \mathcal{N}(\zvec | \mu_0, \sigma^2_0)$\footnote{The mean $\mu_0$ has the same dimension of the vector $\zvec$ and $\sigma^2_0$ as a vector of variances (i.e., all dimensions are independent). We keep this convention through the rest of the paper.}, and a model $p(\avec_i | \zvec) = \mathcal{N}(\phi(\avec_i)| \zvec, \sigma^2(\avec))$ where $\phi$ is a deterministic mapping that project $\avec$ in the same dimension of $\zvec$. BA uses the Bayesian rule to infer $p_{i+1}(\zvec) = p(\zvec | \avec_{i+1}) \propto p(\avec_{i+1} | \zvec)p_i(\zvec)$. This process results in $p(\zvec | \avec_1, \avec_2, \dotsi, \avec_n) = p_0(\zvec)\prod_{i=1}^n p(\avec_i | \zvec)$, which is computable in closed-form for normal distributions. Notice that the latent distribution can be equivalently rewritten as $(\zvec | \avec_1, \avec_2, \dotsi, \avec_n) = p_0(\zvec)\prod_{i=1}^n p(\zvec | \avec_i)$, since, due to the symmetry of the Gaussian distribution, $\mathcal{N}(\phi(\avec_i)| \zvec, \sigma^2(\avec)) = \mathcal{N}(\zvec| \phi(\avec_i), \sigma^2(\avec))$.  Unlike the classic mean estimator, BA builds a probabilistic model of the latent variable based on the conditioning variable. In particular, conditioning variables that carry less information will have a lower impact on predicting the latent variable without lowering its uncertainty. In comparison, context variables that carry more information will tend to decrease model uncertainty and have a higher impact on the prediction. The latent space induced by BA is processed by subsequent nonlinear mappings and is trainable via variational inference.

\section{Model Architecture and Training}
\label{sec:deep-promps}

An essential function of ProMPs is to learn a model of the distribution of motions (trajectories) shown by a human demonstrator. Once trained, the model can be used to sample comparable trajectories to the demonstrations. However, in most applications, one does not want only to replicate the distribution of observed movements but condition them on certain variables such as via-points and context variables.

The primary idea of this work is to use a Bayesian aggregator to model the latent space and to use variational inference to train the model similarly to variational autoencoders (VAEs) \cite{kingma_stochastic_2014,kingma_introduction_2019}. This enables DeepProMPs to exploit the probabilistic model of the latent space to produce a multimodal profile of the epistemic uncertainty and learn the relation between a variable number of conditioning variables and movements. 
\subsection{Model Architecture}
Our model is mainly composed of two elements: the \textsl{encoder}, which consists of a Bayesian aggregator that processes a variable number of via-points and context variables and builds the latent distribution of $\zvec$, and the \textsl{decoder}, a neural network that implements $p(\yvec | \zvec, \xvec)$, where $\xvec$ is a normalized time (or phase), $\zvec$ is the latent representation of the motion, and $\yvec$ is the desired robot configuration. Therefore, the encoder processes two different typologies of inputs: the set of via-points, denoted with $\mathcal{A} = \{\avec_i\}_{i=1}^n$ and the set of context variables, denoted with $\mathcal{C} = \{\cvec_i\}_{i=1}^m$. As described in Section~\ref{sec:background-bca}, the latent distribution is
\begin{align}
    & q(\zvec | \mathcal{A}, \mathcal{C}) \propto p_0(\zvec)\prod_{i=1}^n q(\zvec | \avec_i)\prod_{i=1}^m q(\zvec | \cvec_i),
    \label{eq:variationalposterior}
\end{align}
where $p_0(\zvec) = \mathcal{N}(\zvec | \mathbf{0}, \mathbf{I})$, $q(\zvec | \avec_i) = \mathcal{N}(\zvec | \mu_a(\avec_i), \sigma_a^2(\avec_i))$, and $q(\zvec | \cvec_i) = \mathcal{N}(\zvec | \mu_c(\cvec_i), \sigma_c^2(\cvec_i))$ where $\mu_a, \mu_c, \sigma_a, \sigma_c$ are nonlinear projections encoded by neural networks.
In practice, the mean and the variance of the latent distribution are computed by:
\begin{align*}
    \mu_z(\mathcal{A}, \mathcal{C}) & = \sigma^2_z(\mathcal{A}, \mathcal{C})\left( \sum_{i=1}^n \frac{\mu_a(\avec_i)}{\sigma^2_a(\avec_i)} + \sum_{i=1}^m \frac{\mu_c(\cvec_i)}{\sigma^2_c(\cvec_i)}\right), \\
    \sigma^2_z(\mathcal{A}, \mathcal{C}) & = \left(1 + \sum_{i=1}^n \sigma_a^2(\avec_i)^{-1} + \sum_{i=1}^m \sigma_c^2(\cvec_i)^{-1}\right)^{-1}.
\end{align*}
Notice that we assume a unique pair of functions (i.e., neural networks) to process the mean $\mu_c$ and the variance $\sigma_c^2$ associated with the context variables. However, in some settings, we want to have different kinds of context variables (e.g., raw images, and low-dimensional poses extracted from a motion capture), and we, therefore, employ different neural networks for the different types of input data.
The decoder $p(\yvec | \zvec, \xvec)$ is implemented using a Gaussian distribution, i.e.,
\begin{align*}
    p(\yvec | \zvec, \xvec) = \mathcal{N}\left(\yvec \big| \mu_y(\zvec, \xvec), \sigma^2_y\right),
\end{align*}
where $\mu_y$ is a nonlinear projection encoded with a neural network and $\sigma_y$ is an hyper-parameter. At this point, the probability of a robotic configuration $\yvec$ at time $\xvec$ given a set of via-points and conditioning variables is
\begin{align}
p(\yvec | \xvec, \mathcal{A}, \mathcal{C}) = \int p(\yvec | \xvec, \zvec)p(\zvec | \mathcal{A}, \mathcal{C}) \de \zvec. \label{eq:probability-y}
\end{align}
Unfortunately, \eqref{eq:probability-y} is not solvable in closed form, and we need to resort to \textsl{variational inference} to train the model. 
%{\color{blue} explain what is the architecture of our model primitive.}
% The probilistic MODEL, like the graph. 
% \begin{enumerate}
% \item separate encoders for each type of data \textit{for experiments we share the output head for the latent parameters $\mu, \sigma$ between all encoders}
% \item all same output dimensional size because of shared output heads
% \item we aggregate the latent distribution with Bayesian aggregator
% \item the latent $z \sim N(\mu_{agg}, \sigma_{agg})$
% \item and we pass the $z$ to some decoder $f(z_i, t_i)$
% \end{enumerate}

\subsection{Training the Model via Variational Inference}
Similar to classical variational autoencoders (VAEs) \cite{kingma_stochastic_2014,kingma_introduction_2019}, our model is composed of three entities: model likelihood $p(\yvec | \zvec, \xvec)$, variational posterior $q(\zvec| \mathcal{A}, \mathcal{C})$, and prior distribution $p_0(\zvec)$. The model likelihood can be rewritten using the
reparameterization trick, 
\begin{align}
	\yvec = \mu_y(\zvec, x) + \epsilonvec \sigma_y \quad \text{with}\ \epsilonvec \sim \mathcal{N}(0, I), \label{eq:generation}
\end{align}
to allow gradient training. The evidence lowerbound (ELBO) is
\begin{align}
	\mathcal{L}(\mathcal{A}, \mathcal{C}) = &\underset{\zvec \sim q(\cdot | \mathcal{A}, \mathcal{C})}{\mathbb{E}} \left[\sum_{i=1}^n \log p(\yvec_i | \zvec, x_i)  - \log\frac{q(\zvec | \mathcal{A}, \mathcal{C})}{ p_0(\zvec)}\right]. \nonumber %\label{eq:elbo}
\end{align}
Plugging in the definition in \ref{eq:variationalposterior}, we rewrite the ELBO as
\begin{align}
	\mathcal{L}(\mathcal{A}, \mathcal{C}) = & \underset{\zvec \sim q(\cdot | \mathcal{A}, \mathcal{C})}{\mathbb{E}}\left[\sum_{i=1}^{n} \log p(\yvec_i | \zvec, x_i) \right] \nonumber \\& - \underset{\zvec \sim q(\cdot | \mathcal{A}, \mathcal{C})}{\mathbb{E}}\left[\sum_{i=1}^{n}\log\frac{ q(\zvec | \avec_i)}{p_0(\zvec)} +\sum_{i=1}^{m}\log\frac{q(\zvec | \cvec_i)}{ p_0(\zvec)} \right].\! \nonumber
\end{align}
Notice that the additional KL divergence between the variational posteriors $q(\zvec | \avec_i)$, $q(\zvec | \cvec_i)$ and the prior $p_0(\zvec)$ are meant to satisfy the prior assumption,  i.e., $\mathbb{E}_{\avec}[ p(\zvec | \avec_i)] = \mathbb{E}_{\cvec}[ p(\zvec | \cvec_i)] = p_0(\zvec)$. To train the neural network we maximize the ELBO with respect to the parameters of $\mu_a$, $\sigma_a$, $\mu_c$, $\sigma_c$, and $\mu_y$.
Variational inference suffers from problems like posterior collapse \cite{Bowman2016GeneratingSF} and aliasing from a mismatch of the posterior and prior distributions \cite{bauer_LARS_2019}. The former negatively impacts our movement primitives' ability to replicate the data accurately due to the over-regularizing effects of the KL divergence. The latter can lead to the model generating potentially undesirable demonstrations outside the training distribution.  We address the former problem with KL-annealing and adjusting the weighting of the KL divergence. The latter can be addressed with learnable prior distributions, but we do not consider it in this work. % and we include results demonstrating the adverse effects of prior mismatch on generating unconditional trajectories in {\color{red} Appendix}~\ref{app:theory}.
%COMMENT - maybe move to end?
%Since all the elements contribute linearly with respect to $\mathcal{L}$, the same objective can be optimised by sampling a single element of the trajectories,
%\begin{align}
%	\mathcal{L}(a_i, a_j, c_k) = \mathbb{E}_{\zvec \sim q(\cdot | a_i, c_k)}\left[ \log p(\yvec_i | \zvec, x_i) \right] - \mathbb{E}_{\zvec \sim q(\cdot | A, C)}\left[\frac{\log \hat{f}_A(\zvec | a_j)}{\log p(\zvec)} +\frac{\log \hat{f}_C(\zvec | c_k)}{\log p(\zvec)} \right], \nonumber
%\end{align}
%where $a_i, a_j,$ and $ c_k$ are independently sampled from the same trajectory,  making the computations considerably lighter and compatible with standard stochastic gradient techniques.

\subsection{Operations}\label{sec:operations}
\textbf{Generation of Movements.}
DeepProMPs are a deep-learning variant of ProMPs that can depict complex movement distributions.  To build a trajectory it is sufficient to sample $\zvec$ from the prior distribution $\zvec \sim p_0(\zvec)$ %(which can be learned using LARS \cite{bauer_LARS_2019})
and then use the model in Equation~\ref{eq:generation} to predict $\yvec$ at the appropriate normalized time $\xvec$,
$\yvec = \mu_{y}(\zvec, \xvec)$ without including the noise $\epsilonvec$.
Note that to get a smooth motion, we do not resample $\zvec$ during trajectory generation. %disturb the trajectory with the isotropic noise $\epsilon$.
The variable  $\zvec$ remains constant over the life of a trajectory.

\textbf{Via-Points and Context Conditioning.}
Once our model is learned, we want to use it to predict new trajectories that pass through a set of via-points $\tilde{\mathcal{A}} \equiv \{\tilde{\avec}_i\}$ and is conditioned on a set of external variables $\tilde{\mathcal{C}}$. Notice that one of the two sets can be empty (if they are both empty, we are using our model purely as generative). We use the variational posterior as in Equation~\ref{eq:variationalposterior} to sample $\zvec$, and then we use Equation~\ref{eq:generation} to generate the whole trajectory. However, occasionally, it might be useful to weight some via-points or some external variables differently. Taking inspiration from ProMPs, we rewrite the variational posterior, including importance weights $\omega^{A}_i$, $\omega^{C}_i$,
\begin{align}
	q(\zvec | \tilde{A}, \tilde{C}, \omegavec^A, \omegavec^C) = \prod_{i=1}^{n}q_A(\zvec | \tilde{\avec}_i) ^{\omega^A_i}\prod_{i=1}^{m} q_C(\zvec | \tilde{\cvec}_i)^{\omega^C_i} \label{eq:conditioning}
\end{align}
\begin{figure}
    \centering
    \vspace{0.25cm}
\hspace{-0.3cm}\input{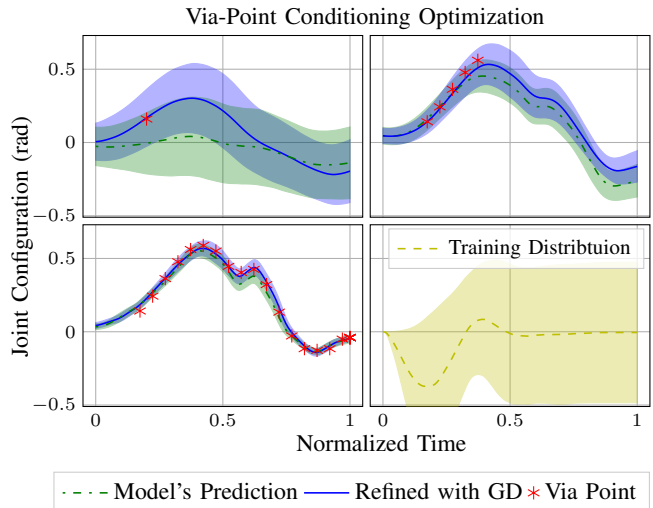}
    \caption{Results of the deployment-time via-point error minimization on one of the joints from our real-robot data. In green is the predicted distribution from our model; in blue is the distribution refined with gradient descent. Benefits are more pronounced with fewer via-points, while more via-points improve the overall quality of the prediction. The bottom-right plot shows the distribution of the training data. }
    \label{fig:via_pt_opt} 
\end{figure}
obtaining a Gaussian distribution again. This reweighting can be useful as the size of via-points conditioning and context conditioning  are usually unbalanced, and the reweighting can help correct this bias. Notice that this importance reweighting is not possible with the mean aggregator. %Furthermore, CNMPs treat context conditioning separately {\color{blue} CNMP treat via point and context as occurring jointly, no? I got this detail wrong in previous submission} from the via-point conditioning, generating, i.e., $\yvec = f(\zvec, \xvec, \cvec)$. Our solution is more flexible since it allows learning both via points and contexts and leaves the choice of which conditioning to use at deployment. 
%{\color{red}
%\paragraph{Addressing Model Mismatch}
%In our experiments, we found that particularly in cases where a single via-point is provided, the model latent distributions can be mismatched and generate distributions that in expectation do not pass through the desired via-point. We found that the predicted latent representation could be optimized to enforce matching the $\mathbb{E}[ \yvec | \zvec]$ to pass through via points with the following optimization loss:
%\begin{equation}
%    \min_{\mu, \log{\sigma}} ||a_{i} - \frac{1}{K}\sum_{i}^{K} \hat{y}(z, x_i))||^{2}_{2}. + ||\sigma^{*} - \text{softplus}(\log{(\sigma)}||^{2}_{2}
%\end{equation}
%where $\hat{y}_(z, x_{i})$ is the prediction of the decoder model at training. At each iteration, we sampled 100 Monte Carlo samples to use the average prediction. Empirically, we found this necessary to help keep variance. We also include matching the variance of the original prediction as we want to preserve the variance. This forces the latent embedding to only match in expectation to the via point. 

%As we directly optimize for the latent distribution parameters, a valid concern is whether the initialization induced yields good initializations for optimization. We conduct an experiment comparing using the prior distribution parameters as the initial starting point $\mu = \textbf{0}$ and $\log{\sigma} = softplus^{-1}(1.0)$. 
%}
%{
%\color{blue} \textbf{Samuele suggested replacement:\\}
The generated motions can arbitrarily violate the desired via-points. To mitigate this issue, we propose to perform a further optimization stage to adjust the distribution of generated motions to minimize the error for specified via-points. This optimization stage, carried out at deployment time, minimizes the distance between the generated trajectories and the target via-points while avoiding variance collapsing, i.e.,
\begin{equation}
    \min_{\tilde{\mu}_z, \tilde{\sigma}_z} \sum_{i=1}^n \Big\| \yvec_i - \frac{1}{k}\sum_{j=i}^k \mu_y(\zvec_j, \xvec_i)\Big\|^{2}_{2} + \left\|\sigma^{*} - \tilde{\sigma}_z \right\|^{2}_{2}, %\label{eq:matching-loss}
    \label{eq:iterativeopt}
\end{equation}
where $\zvec_{j = 1, \dots, k} \sim \mathcal{N}(\tilde{\mu}_z, \tilde{\sigma}_z)$, $k$ is the number of Monte-Carlo samples and $\sigma^*$ is the target variance. We solve \eqref{eq:iterativeopt} by improving the initial solution $\tilde{\mu}_z = \mu_z(\mathcal{A})$ and $\tilde{\sigma}_z = \sigma_z(\mathcal{A})$ via gradient descent (Fig.~\ref{fig:via_pt_opt}). Further details are in Appendix.
%}
\begin{figure}
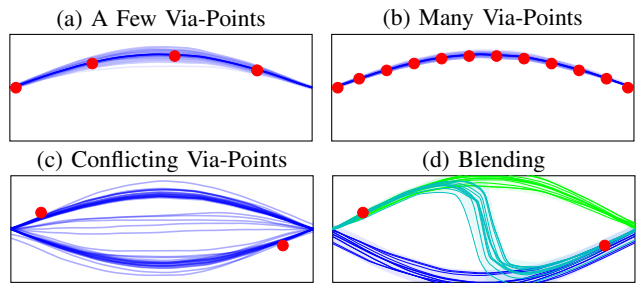

\centering
\vspace{0.3cm}
\input{figures/few-points/sin_4_points_increasing_points.tex}
\input{figures/many-points/sin_12_points_increasing_points.tex}\\ \centering
\hspace{0.15cm}\input{figures/conflict/conflict.tex}
\input{figures/blending/blending.tex}
	\caption{DeepProMPs used in different situations. (a) Distribution of trajectories using $3$ via-points conditioning. (b) With more via-points, the variance of the distribution decreases. (c) With inconsistent via-points, the model chooses to violate one of them staying close to the given dataset and avoiding unseen behavior. Note the ability of our model to generate a bi-modal distribution. (d) Generalization can be enhanced using blending.\label{fig:sine}}\vspace{-1.5em}
\end{figure}
\\\textbf{Blending.}
As proposed in ProMPs, the blending of motions consists of smoothly transitioning from one movement to another. Consider a time-varying weight $\omega(\xvec) \in [0, 1]$ and two movement distributions $q(\zvec | \mathcal{A}_1, \mathcal{C}_1)$ and $q(\zvec | \mathcal{A}_2, \mathcal{C}_2)$; we derive the blending of the two movements
\begin{align}
	q_\text{b}(\zvec|\mathcal{A}_1, \mathcal{A}_2, \mathcal{C}_1, \mathcal{C}_2, \xvec) = q(\zvec | \mathcal{A}_1, \mathcal{C}_1)^{\omega(x)}q(\zvec | \mathcal{A}_2, \mathcal{C}_2)^{1-\omega(x)}. \label{eq:blending}
\end{align}
using the exponential weighting introduced by ProMPs.
Observe that the distribution is now dependent on time. This appears to be a problem, as sampling $\zvec$ at different periods $\xvec$ will result in a jerky motion. Using the reparametrization trick, however, we can resolve this issue by considering the mean $\mu_\text{b}(\mathcal{A}_1, \mathcal{A}_2, \mathcal{C}_1, \mathcal{C}_2, \xvec)$ and standard deviation $\sigma_\text{b}(\mathcal{A}_1, \mathcal{A}_2, \mathcal{C}_1, \mathcal{C}_2,\xvec)$ of the blended latent representation of the motion \eqref{eq:blending}. By first sampling a standard noise $\epsilonvec \sim \mathcal{N}(0, I)$ and then computing 
\begin{align}
	\zvec(\xvec) = \mu_\text{b}(\mathcal{A}_1, \mathcal{A}_2, \mathcal{C}_1, \mathcal{C}_2, \xvec) + \epsilonvec \cdot \sigma_\text{b}(\mathcal{A}_1, \mathcal{A}_2, \mathcal{C}_1, \mathcal{C}_2, \xvec), \nonumber %\label{eq:normal-blending}
\end{align}
 where $\cdot$ is the element-wise product, we obtain a smooth transition of $\zvec$.
%Hower, while blending two movement was essential with ProMPs, since one needed many ProMPs to capture multimodalities, in our application, this kind of blending will be less interesting. However, we can a continuous blending, to overcome the loss of generalit zy described in the generative model. Our model tends to replicate the movement shown with high fiedelty. Very often this is a nice feature, but some time, we want to generalize to completely unseen movement. We can do this by using a set of via points $\tilde{A}$, and giving them a different important during time, i.e., $\omegavec(x)$, allowing a blending of \textsl{infinite} movements,
%\begin{align}
%	q(\zvec | \tilde{A}, \omega, x) = \prod_{i=1}^{|A|} \hat{f}_A(\zvec |\tilde{a}_i)^{\omega_i(x)}.
%\end{align}
%Again, we can use Equation~\ref{eq:normal-blending} to prevent jerky movements.

\textbf{Time Modulation.}
%<<<<<<< HEAD
%Time modulation is a crucial component for movement primitives. During training, time is always normalised between $0$ and $1$, therefore, a time modulation function $\tau(t):[0, T] \to [0, 1]$ can be used to translate the real time $t \in [0, T]$ to the \textsl{phase} space $\xvec \in [0, 1]$. In our implementations, we employ the linear time modulation $\tau = t/T$, where $T$ is the whole duration of the movement. This linear modulation provides a straightforward method for achieving proportional velocity profiles with faster or slower movement. In general, a monotonic $\tau$ can arbitrarily accelerate or decelerate a movement during its execution. If $\tau$ is non-monotonic, it can \textsl{revert} the learned motions. 
%=======
Time modulation is a crucial component for movement primitives. During training, time is always normalised between $0$ and $1$, therefore, a time modulation function $\tau(t):[0, T] \to [0, 1]$ can be used to translate the real-time $t \in [0, T]$ to the \textsl{phase} space $\xvec \in [0, 1]$. In our implementations, we employ the linear time modulation $\tau = t/T$, where $T$ is the whole duration of the movement. This linear modulation provides a straightforward method for achieving proportional velocity profiles with faster or slower movement. In general, a monotonic $\tau$ can arbitrarily accelerate or decelerate a movement during its execution. If $\tau$ is non-monotonic, it can \textsl{revert} the desired motions. 

\textbf{Rhythmic Movements.} Rhythmic movements can be seen as a segment of movement that can be repeated indefinitely many times. Rhythmic movements require these segments to have the same initial and endpoint so that their repetition does not cause jumps.  
To obtain this property, the time-modulation function must have the same position and velocity at both $t=0$ and $t=T$. This effect can be obtained by using the phase $\xvec= [\sin (2 \pi t/T), \cos (2 \pi t/T)]^\intercal$ both during training and deployment. We note this representation can be viewed as positional encoding where we use the behavior of the sinusoidal functions \cite{vaswani_attention_2017}.

\section{Empirical Analysis}
\begin{figure}
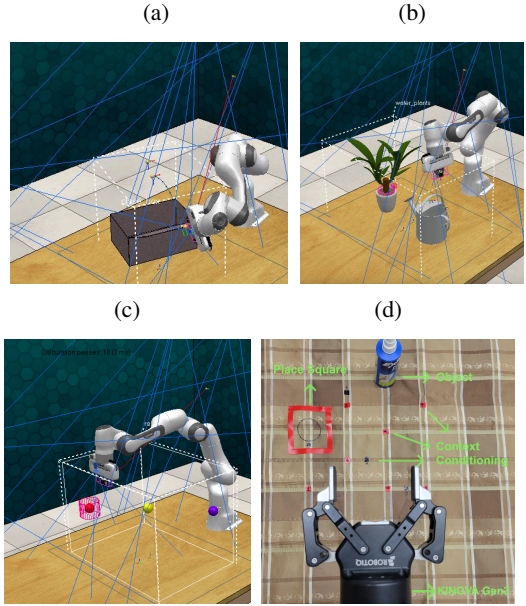

\vspace{0.2cm}
\centering\input{figures/closebox.tex}
\hspace{-0.7cm}
\input{figures/water-plants.tex}
\hspace{-0.4cm}
\input{figures/reaching.tex}
\hspace{-0.45cm}
\input{figures/kinova_pick_n_place}

\caption{(a), (b), (c): Close box,  pour water, and reach from RLBench. (d) A top view of our testing setup: The robot should grab the object and place in the designated square. Positions are encoded with 2D context variables. %\textbf{(d)}: The robot picks up the trash and throws it in the bin.
\label{fig:barret-wam} }
	\vspace{-1em}
\end{figure}
\label{sec:result}
In this section, we conduct several experiments to compare DeepProMPs against alternative motor primitive models. We compare each model's reconstruction capabilities on demonstrations provided from several simulation tasks and a real robotic problem. We choose the Mean Square Error as our metric of choice to highlight how well we can recover demonstrations depending on the conditioning. We also include experiments demonstrating the time-modulation formulation for cyclical tasks on a real robot.

We choose ProMPs, CNMPs, and VAE-CNMPs as our baselines. We include ProMPs because it is the foundational framework that motivates our work. CNMP can be viewed as a version of our model with fixed variance in the latent distribution, but that predicts output variance $\sigma_{y}(\zvec, \xvec)$ as done in previous work \cite{seker_conditional_2019}. VAE-CNMP is a variation of our model without Bayesian aggregation, the via-point, and context variable independence assumption, but still uses the Isotropic Gaussian as the prior in the KL divergence regularization. For both CNMPs and VAE-CNMPs, we use a single encoder that concatenates all inputs (via points and context variables) into a single input to the encoder. We also train a version of CNMPs and VAE-CNMPs with zero-padding to compare whether or not these models are as capable of the optionality of inputs as DeeProMPs. For example, to simulate inputting only via-points with our baselines, the variational encoder's input would be $q(\zvec | \avec,  \cvec^0)$, where $\cvec_0 = \overrightarrow{0}$ or context only as $q(\zvec |\avec^0, \cvec)$ where $\avec^0 = \overrightarrow{0}$ for context only inputs. We refer to these versions as CNMPs (Indep) and VAE-CNMPs (Indep) in our results. 

We conduct experiments with demonstrations collected in three simulated tasks from RLBench \cite{james_rlbench_2020}, and two sets of demonstrations performed on physical robotic manipulators. The three simulation tasks we use are: (1) \texttt{Reach}, a task where the manipulator goes to a designated target, (2) \texttt{Close Box}, a task where the manipulator closes a container, and (3) \texttt{Pour Water}, a task where the robot must pick up and pour the water from a container. Our real robot experiments include demonstrations of a pick-and-place task where a Kinova Gen-3 lite moves a designated object to a specified location. We refer to the Kinova pick-and-place task as the \texttt{Kinova} task in this section. We use the previous four tasks (three simulation and one real robot) to compare reconstruction performance. The second robot task uses a Barrett WAM$^\circledR$ Arm to shake a container which we use to verify training and execution of cyclical tasks. The set-up for the latter simulation tasks and both real robot tasks are featured in Figure~\ref{fig:barret-wam}.

In our reconstruction experiments, we vary the number of trajectories used in each experiment. Our choices in the simulation task were motivated by preliminary experiment results and the complexity of the task. In the real robotics experiments, we used kinesthetic demonstration to collect the data. Respectively, we use 500 training examples for the Reach task, 1000 demonstrations for the Close Box and Pour Water task, and 100 demonstrations for the Kinova task. For all simulation tasks, we select the best models using a set of 100 validation examples and report results on 200 test examples. In the Kinova experiments, we report the best validation results using 10 demonstrations for all models. 

For each reconstruction experiment, we train all models to take via-points, low-dimensional context variables, and images as inputs. The dimension of the low-dimensional context varies for each task. In the Reach task, this is the target location (3 dimensions); in Close Box, this includes the angle of the box’s lid and box location (54 dimensions); for Water Plants, this is the location of the watering can and object to be watered (84 dimensions). In the Kinova experiments, this is the normalized pixel location of objects (2 dimensions) and one-hot encoding for the sub-task (pick or place). In the simulation tasks, each trajectory is associated with five images of the scene. We always use these images together at deployment. For our Kinova data, we take a picture of the object to be moved and augment this with three additional pictures that contain markers on the desired object. We found that this mitigated issues with the limited training data used. Further investigation in addressing this issue for deep motor primitives is potential future work, but out-of-scope of this paper which focuses on proposing the DeepProMPs framework generally.

\textbf{Task Reconstruction.} We evaluate ProMPs, DeepProMPs, CNMPs, VAE-CNMPs, CNMP (Indep), and VAE-CNMP (Indep). We used the Adam optimizer \cite{kingma_adam_2014} with all models using a learning rate of 1e-4. We train each model for 8000 epochs in our simulation experiments and 10000 epochs for our Kinova experiments, while always using mini-batches of size 16. All models use two hidden layer multi-layer perceptrons with ReLU activations for vector inputs (via-points and low-dimensional context variables). Each layer has 128 neurons. All models have a single decoder with two hidden layers each with 128 neurons that take in the latent variable and time-modulation variable as input. For images, we use a Resnet-50 architecture \cite{he2015Residualconnections} to embed images in the latent space. We aggregate the representation for all images associated with a trajectory using a mean operation. Both CNMP and VAE-CNMP concatenate all context variable types to each via point as inputs. When using DeepProMP, all encoders share a single affine head to produce latent distribution parameters. We use latent dimensions of 16 for Reach and Kinova and 32 for Close Box and Pour water for all models. During training, we sub-sample combinations of input data for each model. For DeepPromp, CNMP (Indep), and VAE-CNMP (Indep) this includes an added step of choosing combinations of via-point, low-dimensional, and image context variables as inputs. For example, training with via-points and images on one update and via-points only on another. Both CNMP and VAE-CNMP models always receive a via-point and context variable combination as input because of the concatenation operation. We report results using models with the best validation error for each metric reported and train 5 of each model which results are averaged over. The one exception is ProMP, which has a deterministic solution and we instead bootstrap results to average results. 

\begin{figure}
    \centering
    \vspace{0.5cm}
    \hspace{2.5cm}
    %\includegraphics[%natwidth=296mm, natheight=319mm, 
    %width=0.5\textwidth]{figures/radar_plots/radars_plots.pdf}
    \includegraphics[%natwidth=296mm, natheight=319mm, 
    width=0.5\textwidth]{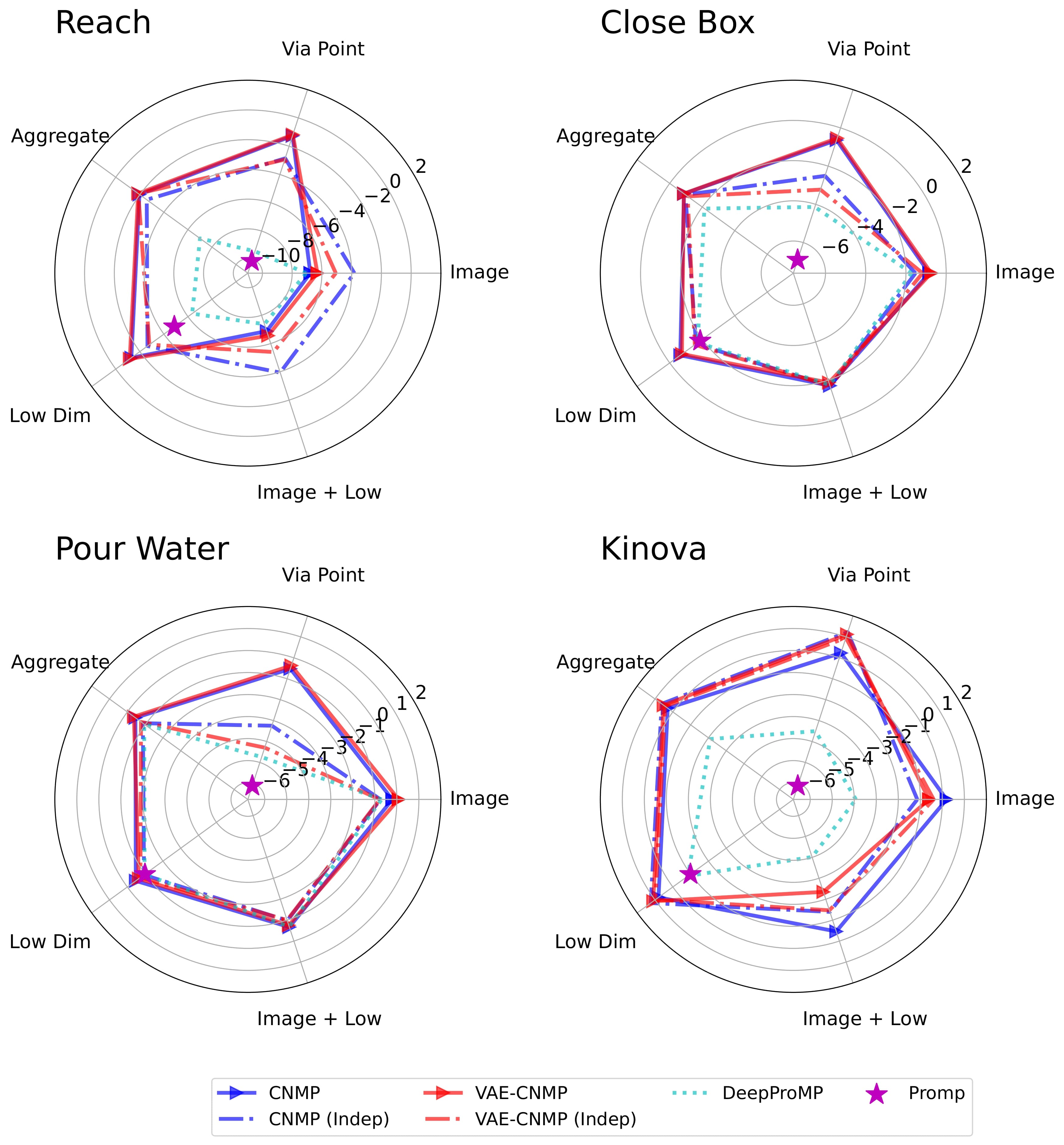}
    \caption{
    Radar plots comparing reconstruction performance across different motor primitive models. Smaller circles are indicative of better performance across potential data types. We consider conditioning on images, low-dimensional context variables, the combination of both, the full trajectory as via points and the average across the four former types of inputs. Measurements are in log scale of the mean square error. 
    }
    %Visualization of the reconstruction mean squared error (log scale) on the test set for different types of conditioning and via point combinations ({\color{red} explain here or in the text what "via point", "aggregate", "low dim" etc stand for}). Notice that ProMPs are tested only for low dimensional context conditioning and via point conditioning settings which we mark with stars. {\color{red} TODO}}
    \label{fig:radars}
\end{figure}

We report the average performances with radar plots in Figure~\ref{fig:radars}. Smaller circles imply better performance across all metrics. We show the log-scale Mean Square Error of reconstructing demonstrations given different combinations of inputs. For CNMP and VAE-CNMP variants we achieve this with zero padding as previously described to mask out the excluded inputs. \texttt{Via Point} refers to reconstructing the trajectory without any context variables, \texttt{Image} uses only images, \texttt{Low Dim} uses only the low dimensional context variable, and \texttt{Image + Low} uses both context variables. We also include \texttt{Aggregate} which is the average performance across the previous four combinations of inputs. We note that for each context variable input, we include the initial robot position as well.

In via-point conditioning, we see that ProMP is superior across tasks. This result is expected because ProMP's analytic conditioning preserves all the information in the trajectory. Due to compressing the representation in a latent space, all deep models lose some information reconstructing the model. We note that results could be improved with our deployment optimization technique which was not used in these experiments.

Interestingly, we see that the zero padding training is crucial for improving baselines' ability to do via-point conditioning across tasks but is still worse than DeepProMPs. This suggests that modularizing encoder representations to data types is better than using a single module. One advantage DeepProMPs has is great capacity with independent encoders for each data type. We trained several models with double the neurons reported in the experiments for both independent CNMP variants (256 neurons vs 128 neurons per layer), but these yielded inconsistent performance gains and losses across metrics despite having a comparable number of parameters. This is a sensible result because all models are bottle-necked by the dimensions of the latent space despite the size of the encoders.

Our results suggest that despite the independence assumptions in DeepProMP, the model provides competitive results with different input combinations or else beats other methods. We surprisingly find ProMP produces good results on low dimension reconstruction across tasks but is not as easy to extend to image data. We find that despite the independence assumption, our model still performs comparably to CNMP and VAE-CNMP, which are trained to always see all context information during training. In certain cases, the zero-pad training seems to worsen performance in scenarios we would not expect for baseline methods (see Reach results using only images). Our conclusion is that even when baselines are trained with techniques similar to DeepProMPs they come with trade-offs across conditioning choices whereas DeepProMPs do not.

\begin{figure}
	\centering
\input{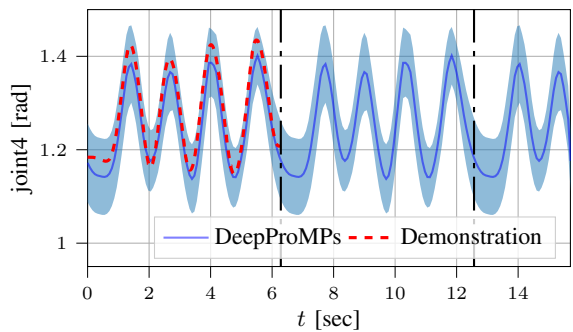}
\caption{Demonstrated motion and learned cyclical behavior.}
\label{fig:cyclicalmotion}
\vspace{-1.5em}
\end{figure}

\textbf{Real-Robot Make Mojito.}
This experiment aims to highlight the ability of DeepProMPs to learn and replay cyclical behaviors. To this end, we collect a small data set of cyclical shaking motions and train  DeepProMPs using the sinusoidal phase modulation proposed in Section~\ref{sec:deep-promps}. In Figure~\ref{fig:cyclicalmotion}, we report the most salient dimension of the learned motion, corresponding to the robotic manipulator's second joint. %Even though the demonstration is not perfect (as its starting point does not match with its end point)
Our model generalized the provided demonstration to perform it in a smooth, cyclic pattern. We find that our model realizes a smooth transition even if the gap between the demonstration's starting point and the endpoint is large. %Further results and details are provided in Appendix~\ref{app:experiment}.

\section{Limitations}

Although DeepProMPs is a promising framework, they do not come without limitations and trade-offs. Deep neural networks and variational inferences add sources of approximation, which were not present in original ProMPs. Many quantities (like via-point and context conditioning) that cannot be found in closed form require sub-sampling training schemes before deployment to achieve these properties. 
Using an approximated posterior introduces a bias in the parameter estimation. However, this is often the price for obtaining a more flexible tool. As we chose to use an isotropic Gaussian for our latent distributions, covariant relations between latent dimensions may not be captured by our model. This could limit the expressiveness of our learned representations. Our model also requires Monte-Carlo sampling in order to estimate the trajectory distribution statistics, which can slow inference time. 
This Monte Carlo sampling is used in our proposed iterative optimization to improve via-point conditioning. There is potential room to study alternative optimization approaches to perform this trajectory post-processing step.

\section{Conclusion}

This paper proposes Deep Probabilistic Motor Primitives as a deep learning variant of ProMPs. Our model is capable of both via-point and context conditioning independent of each. DeepProMPs are more robust to this feature compared to the baseline method trained with zero padding to achieve the same behavior. Our model can also blend motor skills together because of the Bayesian aggregation we incorporated in our model. We demonstrated DeepProMPs' capability of rhythmic motion modulation, which has otherwise been ignored in previous works. Our work is a step towards improving deep motor primitive models for applications in robotics. Future work could include incorporating other complex data types in deep motor primitives like text, consider alternative latent distribution choices to the Gaussian distribution in our Bayesian aggregator, and deploying DeepProMPs in downstream robotic applications.

%This paper proposes a novel approach to deep-sets designed explicitly for movement primitive representation. Our approach can be seen as a generalization of the average aggregator  used in state-of-the-art implementations (such as CNMPs). In addition, state-of-the-art approaches do not consider many desirable properties of movement primitives altogether, such as blending, temporal modulation, or the ability to provide rhythmic movements. In this paper, instead, we address this issue by providing a solution inspired by classic ProMPs.
%We measure the effectiveness of our approach compared to ProMPs, CNMPs and VAE-CNMPs, showing that our approach allows modeling multimodal datasets of movements (like CNMPs) while maintaining the advantageous properties of ProMPs. In particular, our method exhibits higher precision in simulated environments and successfully solves a trash-cleaning task, learning from twenty eight real-world demonstrations. 

\bibliographystyle{IEEEtranS}
\bibliography{bibliography}

\addtolength{\textheight}{-12cm}   % This command serves to balance the column lengths

\section*{APPENDIX}
In Section~\ref{sec:operations}, we explained that the trajectories predicted by the model are not necessarily compliant with the desired via-point. For this reason, we provide a further optimization step at deployment time, by applying gradient descent on the parameters of the latent distribution (mean and variance) to find a distribution of movements that best matches the desired via points. 
\begin{figure}[h]
%\vspace{0.2cm}
\input{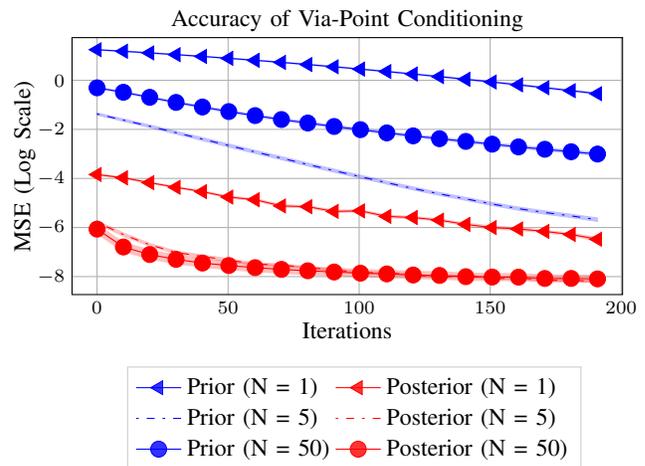}
\caption{Post-processing optimization results for generating trajectories that pass through via-points. We compare $N=1, 5, 50$ context points. For each setting we use fifty different examples and average performance over 200 iterations. We plot the standard-error bars to give confidence in the mean performance. Posterior is the predicted parameters with the variational posterior and Prior uses Isotropic Gaussian parameters and initialization. %We find that initializing gradient descent from the model's prediction (posterior) allows finding a better solution faster while using the prior $p_0(\zvec)$ as starting solution is inefficient. We compare $N=1, 5, 50$ via point conditioning on real-robot data.  
\label{fig:via-point-opt-stuff}
}
\end{figure}

At this point one can question the role of model prediction, since the parameters of the distribution can be found at deployment time. The model prediction serves as a good initialization for the gradient descent optimization. In the experiment in Fig~\ref{fig:via_pt_opt}~and~\ref{fig:via-point-opt-stuff}, we trained DeepProMPs on the Close Box task. We then queried our model to satisfy $N=\{1, 5, 50\}$ via points. For a small number of via points (e.g., 1 and 5) our model does not predict good trajectories, as it is also possible to observe in Fig.~\ref{fig:via_pt_opt}. Gradient descent allows refining such prediction. If we initialize gradient descent with the prior distribution $\mathcal{N}(0, I)$, the optimization procedure will need many training steps to find trajectories compliant with the via points. But when initialized with the distribution predicted by the model, the gradient optimization can find trajectories that meet the via-points with accuracy ranging $10^{-6}-10^{-8}$ radiants in only $200$ steps, as shown in Fig.~\ref{fig:via-point-opt-stuff}.

%\subsection{{\color{red}Anything Else?}}

% \section*{ACKNOWLEDGMENT}

% We thank the all mighty all star for their guidance. 

% %%%%%%%%%%%%%%%%%%%%%%%%%%%%%%%%%%%%%%%%%%%%%%%%%%%%%%%%%%%%%%%%%%%%%%%%%%%%%%%%

% References are important to the reader; therefore, each citation must be complete and correct. If at all possible, references should be commonly available publications.

\end{document}